\pdfoutput=1

\documentclass[11pt]{article}

\usepackage{acl}

\usepackage{times}
\usepackage{latexsym}
\usepackage{graphicx}

\usepackage{microtype}
\usepackage{url}
\usepackage{amsfonts}
\usepackage{amsmath}
\usepackage{amssymb}
\usepackage[font=footnotesize]{subfig}
\usepackage{float}
\usepackage{adjustbox}

\usepackage[T1]{fontenc}

\usepackage[utf8]{inputenc}

\usepackage{microtype}

\usepackage{inconsolata}
\usepackage{multicol}
\usepackage{multirow}
\usepackage{makecell}

\title{Don't Blame the Data, Blame the Model: Understanding Noise and  Bias When Learning from Subjective Annotations}

\author{
Abhishek Anand$^1$, Negar Mokhberian$^{1,2}$, Prathyusha Naresh Kumar$^1$, Anweasha Saha$^1$\\
\textbf{Zihao He$^{1,2}$, Ashwin Rao$^{1,2}$, Fred Morstatter$^2$, Kristina Lerman$^2$} \\
$^1$Department of Computer Science, University of Southern California\\
$^2$Information Sciences Institute, University of Southern California\\
\texttt{\{anandabh, nmokhber, nareshku, anweasha, zihaoh, mohanrao\}@usc.edu} \\ 
\texttt{\{lerman, fredmors\}@isi.edu}
\\
}

\begin{document}
\maketitle
\begin{abstract}
Researchers have raised awareness about the harms of aggregating labels especially in subjective tasks that naturally contain disagreements among human annotators. In this work we show that models that are only provided aggregated labels show low confidence on high-disagreement data instances. While previous studies consider such instances as mislabeled, we argue that the reason the high-disagreement text instances have been \textit{hard-to-learn} is that the conventional aggregated models underperform in extracting useful signals from subjective tasks. Inspired by recent studies demonstrating the effectiveness of learning from raw annotations, we investigate classifying using Multiple Ground Truth (Multi-GT) approaches. Our experiments show an improvement of confidence for the high-disagreement instances\footnote{Our code and data are publicly available at \href{https://github.com/abhishekanand1710/NoiseAndBias}{GitHub}.}. 
\end{abstract}


\section{Introduction}
\textcolor{red}{[Warning: This paper may contain offensive content.]}

Datasets labeled by human annotators play a critical role in many supervised Natural Language Processing (NLP) tasks~\cite{paullada2021data}.
However, as the volume of such data has grown, it has become difficult to manually assess data quality. Recognizing this challenge, recent efforts have proposed automated strategies for evaluating annotated datasets, specifically targeting the identification of noisy and hard-to-learn data instances \cite{swayamdipta-etal-2020-dataset}. 

Existing methods for automatically gauging sample quality often rely on aggregated labels, such as a majority vote \cite{swayamdipta-etal-2020-dataset}, but disagreements among annotations for data items are widespread \cite{plank-2022-problem}. Some of these discrepancies arise from human labeling errors \cite{mokhberian2022noise}, however, a growing body of research highlights that annotator differences in subjective tasks introduce bias in annotations, particularly in sensitive domains like hate speech recognition \cite{plank-etal-2014-linguistically, aroyo2015truth, pavlick-kwiatkowski-2019-inherent, sap-etal-2022-annotators}. Therefore, a single ground truth for each data instance may lead to potential oversights in capturing nuanced perspectives from different annotators.

In this paper, we leverage Data Maps \cite{swayamdipta-etal-2020-dataset}, an automated data evaluation strategy, to understand the relation between noise and bias in annotated datasets. Data Maps define two intuitive measures for each data item: the model's confidence in predicting the true class and the variability of this confidence across epochs. \citet{swayamdipta-etal-2020-dataset} have shown that lower model confidences correlate with higher chances of mislabeling for corresponding samples. Firstly, based on the assumption that a single correct label exists for a given example, we investigate an initial research question:


\textbf{RQ1}: \emph{Is there any correlation between human disagreement on instances and model’s uncertainty/confidence for classifying the instance to aggregated ground truth?}

\citet{swayamdipta-etal-2020-dataset} has briefly studied the relationship between intrinsic uncertainty and the training dynamic measures. Their findings reveal a correlation between human disagreement and the model's uncertainty in a natural language inference dataset. We explore this correlation in the context of toxicity detection in social media texts using three different datasets. Our findings reveal a significant correlation between human label agreement and model confidence, with confidence decreasing as disagreements among annotators increase. Specifically, \textbf{s}ingle \textbf{g}round \textbf{t}ruth (Single-GT) models (see \S \ref{sec:single_gt} for details) exhibit lower confidence for high-disagreement samples, potentially due to the subjectivity of those instances. These observations from \textit{RQ1} motivate the exploration of \textbf{m}ultiple \textbf{g}round \textbf{t}ruths (Multi-GT) or Multi-GT models (details in \S \ref{sec:multi})
that can infer based on multiple perspectives \cite{davani-etal-2022-dealing, gordon2022jury, weerasooriya-etal-2023-disagreement, mokhberian2023capturing} as an alternative to Single-GT models. 

As far as we are aware, there is limited existing research that has examined the training dynamics of non-aggregated annotations. Therefore, we adapt the Data Maps definition to Multi-GT models and empirically address our second research question:

\textbf{RQ2:}. \emph{Does learning from raw annotations enhance the model's confidence for the high-disagreement instances?}

When using Multi-GT models, we identify improved confidence among minority votes for samples characterized by substantial annotation disagreements. 

Our analysis in this paper demonstrates that samples receiving low confidence in Single-GT models are not inherently unusable. Furthermore, employing Multi-GT models for subjective tasks yields improved confidence for certain raw annotations associated with high-disagreement samples.

\section{Related Work}
\paragraph{Uncertainty in Machine Learning}
In the realm of uncertainty estimation and dataset evaluation, several studies have paved the way for understanding the dynamics of model training. \citet{JMLR:v15:srivastava14a} introduce dropout-based uncertainty estimates, showcasing a positive relationship between training dynamics and dropout measures. The Data Maps approach \cite{swayamdipta-etal-2020-dataset} leverages this knowledge to establish the credibility of the proposed training dynamics measures and their relationship with uncertainty. Other works \cite{lakshminarayanan2017simple, gustafsson2020evaluating, ovadia2019trust} collectively support the notion that deep ensembles provide well-calibrated uncertainty estimates, laying the groundwork for our exploration of training dynamics measures and their correlation with uncertainty. \citet{fort2020deep} sheds light on diversity trade-offs in ensembles, offering insights into the cost-effectiveness of using ensembles of training checkpoints. \citet{chen2017checkpoint} advocates for ensembles of training checkpoints as a more economical alternative with certain advantages. The work by \cite{xing2018walk} on loss landscapes provides additional perspectives on the optimization process during training, complementing the understanding gained from training dynamics.

\citet{toneva2019empirical} and \cite{NEURIPS2020_2f3bbb97} , along with \cite{krymolowski-2002-distinguishing} , address catastrophic forgetting, providing approaches to analyze data instances. \citet{bras2020adversarial} introduces AFLite, an adversarial filtering algorithm, advocating for the removal of "easy" instances. \citet{chang2018active} proposes active bias for training more accurate neural networks, aligning with the broader discussion on active learning methods presented in \cite{peris-casacuberta-2018-active, p-v-s-meyer-2019-data}. \citet{maddox2019simple} propose a technique for representing uncertainty in deep learning models utilizing Stochastic Weight Averaging to track a weighted average of neural network weights. \citet{mishra2020dqi} explores creating better datasets, resonating with the theme of dataset enhancement in the context of active learning methods. Influence functions \cite{koh2020understanding}, forgetting events \cite{toneva2019empirical}, cross-validation \cite{chen2019understanding}, Shapley values \cite{ghorbani2019data}, and the area-under-margin metric \cite{pleiss2020identifying} contribute to the discussion on data error detection and instance scoring.

\paragraph{Multiple Perspectives}
The challenge of human label variation due to annotator perspective biases emphasizes the impact on data quality, modeling, and evaluation stages \cite{plank-2022-problem}. Such biases can be propagated to the language models that are pretrained on data from different perspectives \cite{santurkar2023whose, durmus2023towards, he2024whose}.
This resonates with our exploration of model confidence and the drawbacks of aggregating labels in subjective tasks. The call for Multi-GT designs aligns with our goal of understanding noise and bias in raw annotations.

\citet{10.1145/3580494} provides insights into mitigating bias in toxic speech detection, reflecting the awareness raised by researchers about the harms of aggregating labels, especially in tasks involving disagreements among human annotators. It contributes relevant perspectives for enhancing the robustness and fairness of models in the context of subjective tasks.

Prior research has introduced models aimed at directly learning from annotation disagreements in subjective tasks. Two primary approaches have been proposed in this regard. The first approach treats the "ground truth" as the distribution encompassing all labels that a population of annotators could generate \cite{peterson2019human, uma2020case}. The second approach involves learning from the hard labels assigned by individual annotators \cite{davani-etal-2022-dealing, weerasooriya-etal-2023-disagreement, mokhberian2023capturing}.

While preceding studies have made significant strides in uncertainty estimation and dataset evaluation, our work adopts a novel perspective by questioning the effectiveness of aggregated models in identifying mislabeled samples. The definition of confidence used in this study and the Data Maps approach deviates from conventional usage in other fields, where confidence is typically assessed based on the predicted label. Alternative definitions and interpretations of confidence are present in certain core machine learning papers. The shift toward Multi-GT approaches and the exploration of diverse perspectives contribute to a more nuanced understanding of noise and bias within annotated datasets. \citet{wang-plank-2023-actor} suggests innovative uncertainty measures derived from Multi-GT models for integration into an Active Learning pipeline, aiming to decrease the budget required for item-annotator labeling. In contrast, our approach diverges as we focus on exploring training dynamics to capture noise in Multi-GT models.

\section{Datasets}
\label{sec:data}
In this section we introduce the three datasets studied in this paper. Statistics of the datasets are presented in Table \ref{tab:dataset}.

\begin{table}[ht]
\centering
\addtolength{\tabcolsep}{-2.0pt}
\begin{tabular}{lccc}
\Xhline{1.0pt}
\multicolumn{1}{c}{}                                                   & $\mathcal{D}_{\textsl{SI}}$ & $\mathcal{D}_{\textsl{MHS}}$ & $\mathcal{D}_{\textsl{MDA}}$ \\ \Xhline{1.0pt}
\# unique texts                                                        & 45,318  & 39,565  & 10,440 \\ \hline
\# labels                                                              & 2       & 3       & 2      \\ \hline
\# annotators                                                          & 307     & 7,912   & 819    \\ \hline
\begin{tabular}[c]{@{}l@{}}\# annotations \\ per text\end{tabular}     & 3.2$\pm$1.2 & 2.3$\pm$1.0 & 5      \\ \hline
\begin{tabular}[c]{@{}l@{}}\# annotations\\ per annotator\end{tabular} & 479$\pm$830 & 17$\pm$4    & 64$\pm$139 \\ \Xhline{1.0pt}
\end{tabular}
\addtolength{\tabcolsep}{-2.0pt}
\caption{The statistics for dataests introduced in \S \ref{sec:data}}
\label{tab:dataset}
\end{table}

\paragraph{The social bias inference corpus ($\mathcal{D}_{\textsl{SI}}$)} contains 45K posts from online social platforms such as Reddit, Twitter, and hate sites \cite{sap-etal-2020-social}. The dataset includes structured annotations of social media posts with respect to offensiveness, intent to offend, lewdness, group implications, targeted group, implied statement, and in-group language. Following \citet{weerasooriya-etal-2023-disagreement} we only consider the labels from “intent to offend” for each data item. 

\paragraph{The measuring hate speech corpus ($\mathcal{D}_{\textsl{MHS}}$)} consists of 39,565 social media posts spanning YouTube, Reddit, and Twitter, manually annotated by 7,912 Amazon Mechanical Turk annotators from United States \cite{kennedy2020constructing, sachdeva-etal-2022-measuring}. Annotations for each text sample include evaluating the intensity of 10 distinct hate speech labels, encompassing sentiment, disrespect, insult, humiliation, inferior status, violence, dehumanization, genocide, attack or defense, and hate speech. The labels are aggregated across all annotations for a given text using Rasch measurement theory \cite{rasch1960studies}, resulting in a continuous hate speech score, where higher values denote increased offensiveness. This score is discretized into three labels: above +0.5 for hate speech, below -1.0 for supportive speech, and between -1.0 and +0.5 for neutral or ambiguous speech. We use these aggregated labels for Single-GT model. Furthermore, we incorporate each individual annotator's hate speech label as their specific annotation for Multi-GT model. Both the aggregated and non-aggregated target columns represent a multi-class classification task with 3 labels - supportive, neutral, or hate speech. 

\paragraph{The Multi-Domain Agreement dataset ($\mathcal{D}_{\textsl{MDA}}$)} has been created for studying offensive language detection \cite{leonardelli-etal-2021-agreeing}. It comprises approximately 400K English tweets from three topics: Covid-19, US Presidential elections, and the Black Lives Matter movement. Each tweet has been annotated for being offensive or not by 5 US native speakers using Amazon Mechanical Turk, resulting in a total of 10,753 annotated tweets. The tweets have been analysed further in \citet{leonardelli-etal-2021-agreeing} regarding level of annotator agreement: unanimous, mild, and low.

\begin{figure*}[th]
    \centering
    \includegraphics[width=0.32\linewidth]{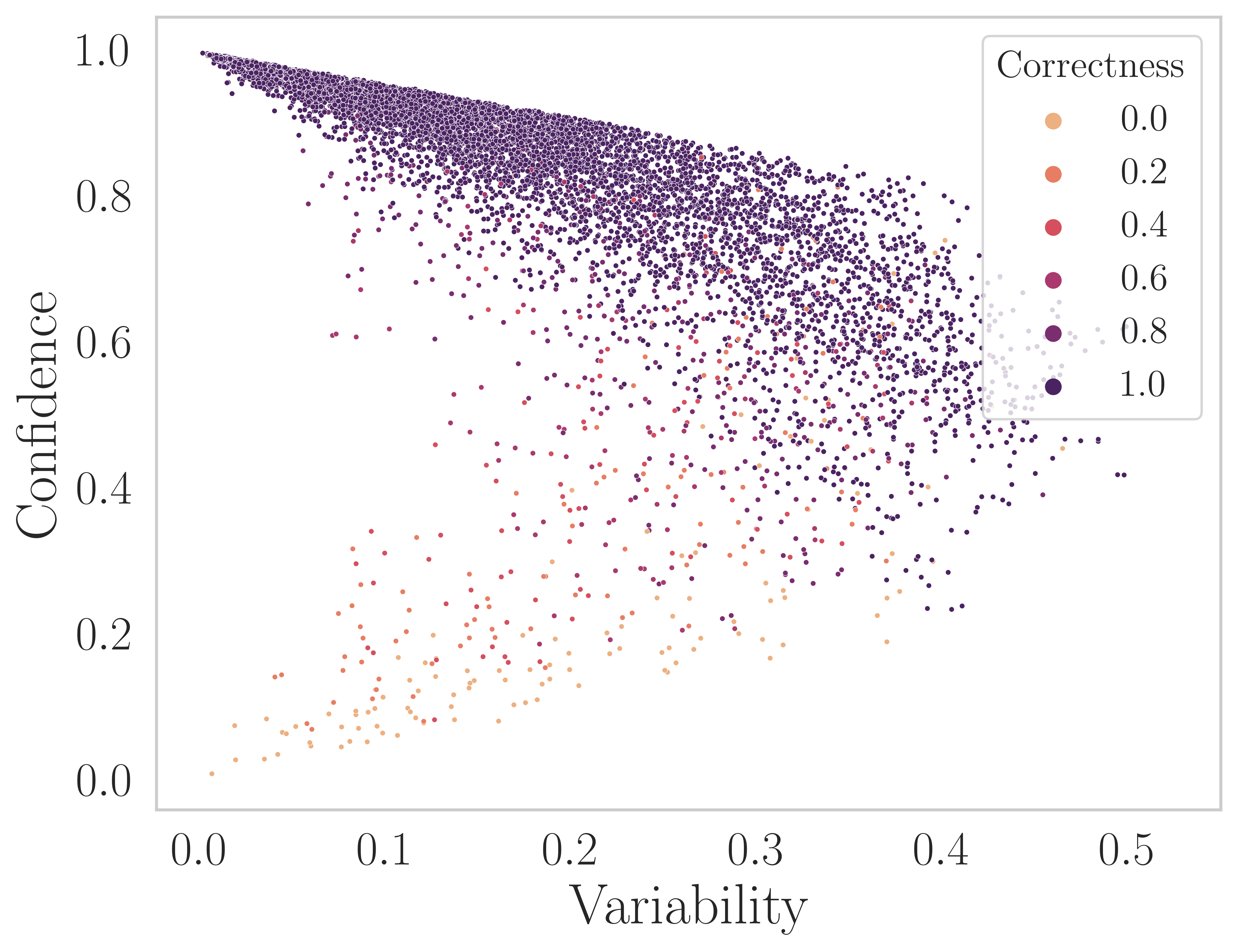}
    \includegraphics[width=0.32\linewidth]{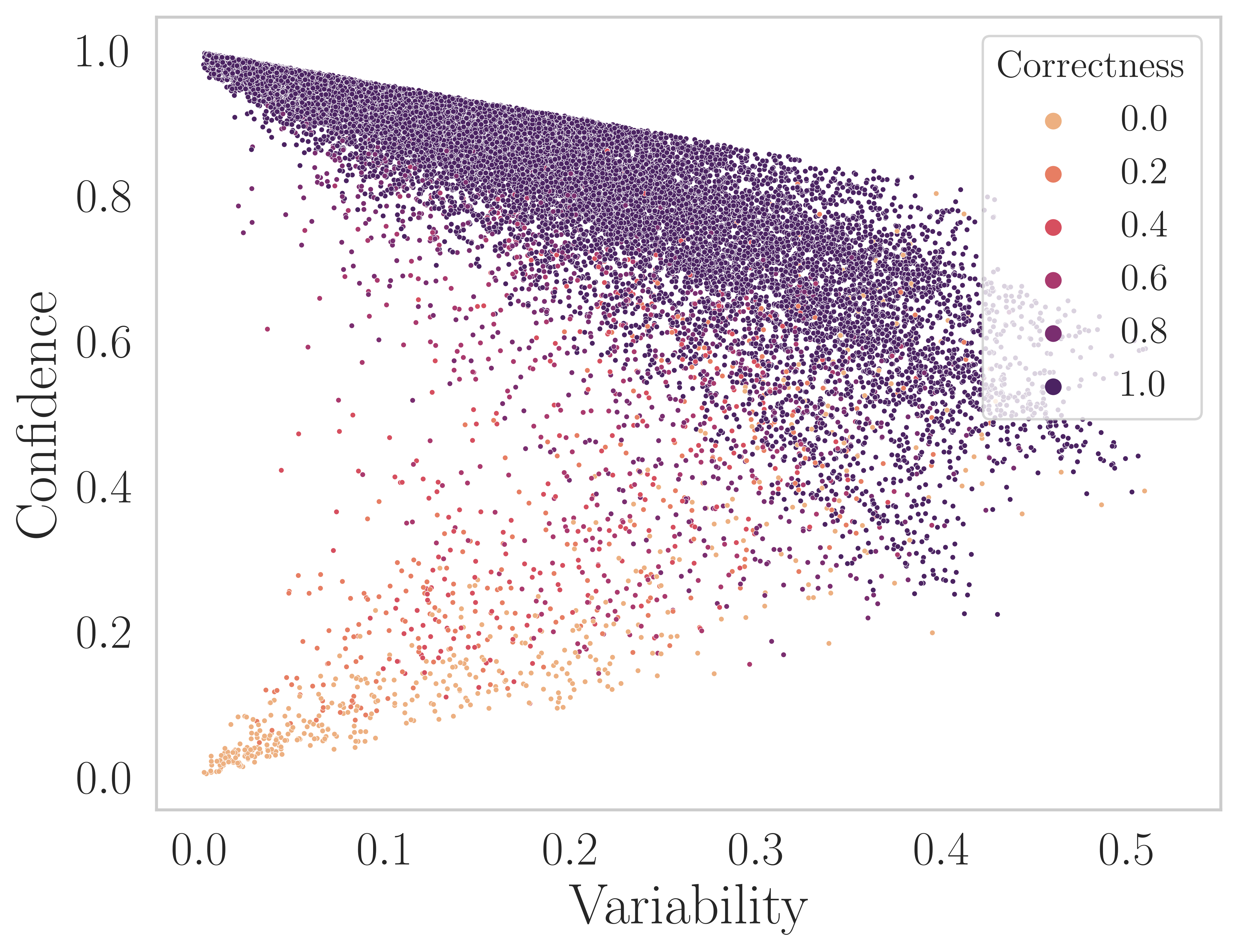}
    \includegraphics[width=0.32\linewidth]{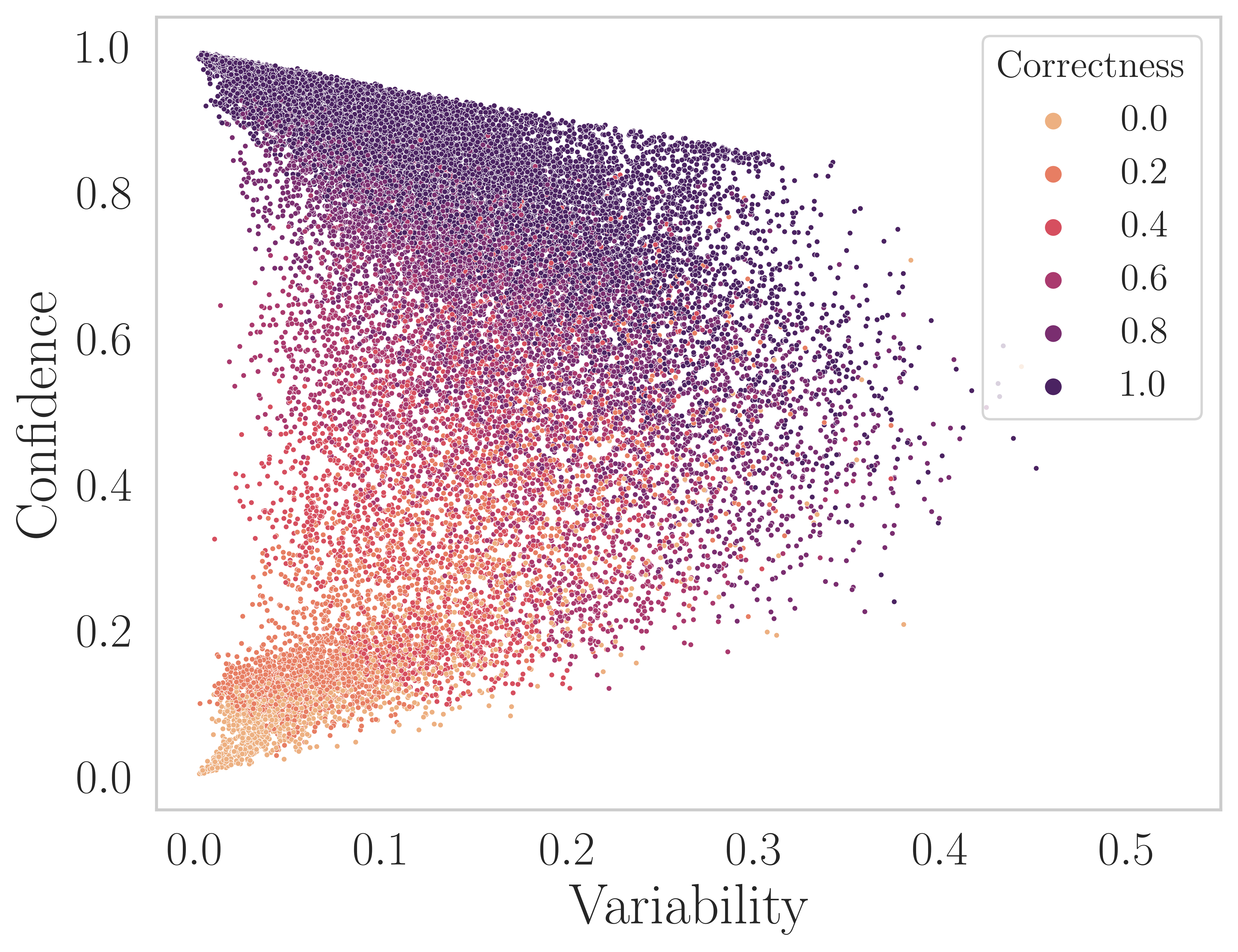}
    \caption{Dataset Cartography map for Single-GT model on $\mathcal{D}_{\textsl{MDA}}$ (left), $\mathcal{D}_{\textsl{SI}}$ (center) and $\mathcal{D}_{\textsl{MHS}}$ (right). The x-axis shows variability and y-axis, the confidence. Further, the points are color-graded by correctness (probability the trained model assigns this data point to the ground truth label in its prediction). Samples in the top left corner with high confidence and low variability are easy for the model to learn, whereas sample that are in the lower left corner with low confidence and low variability are difficult.}
    \label{fig:single_gt_cartography}
\end{figure*}

\section{RQ1: Is there correlation between human disagreement and model’s uncertainty/confidence?}
\label{RQ1}

\subsection{Methods}
This section outlines the approaches employed to address \textit{RQ1}. We compute the agreement level in the human labels directly based on the annotations available in each dataset, with detailed explanations provided in \S \ref{sec:disagreement}. Subsequently, we investigate whether the classifiers' confidence in data items correlates with the level of annotator agreement. We utilize a conventional supervised text classification model, as elucidated in \S \ref{sec:single_gt}, and examine the training dynamics during defined epochs, outlined in \S \ref{sec:train_dyn}.

\subsubsection{Annotator Agreement Level}
\label{sec:disagreement}
The annotator agreement level is defined as the proportion of annotations that align with the majority vote for a specific text sample. This metric, introduced by \cite{Wan_Kim_Kang_2023}, provides insights into the degree of consensus among annotators regarding the majority label assigned to a given sample.

\subsubsection{Single-GT Models}
\label{sec:single_gt}
The conventional text classification model predicts the aggregated label for each instance. Text embeddings from transformer-based encoders are fed into a feed-forward classification layer which performs a linear projection layer to predict the majority label.

\subsubsection{Data Maps}
\label{sec:train_dyn}
We adhere to the definitions outlined in \citet{swayamdipta-etal-2020-dataset} to quantify the qualities of data instances automated by training classification models.

\paragraph{Confidence} is defined as the mean class probability for each data item's gold label across all epochs. The confidence is tied to the evolution of class probabilities during the training process, offering insights into the model's certainty or consistency in predicting gold labels for each data item.

\paragraph{Variability} is defined as the standard deviation of class probability for each data item's gold label across all epochs and measures the extent to which they change across different training epochs. It indicates the degree of fluctuation or stability in the model's predictions over time.

\citet{swayamdipta-etal-2020-dataset} find that the simultaneous occurrence of low confidence and low variability correlates well with an item having an incorrect label.

\begin{figure*}[th]
    \centering
    \includegraphics[width=0.32\linewidth]{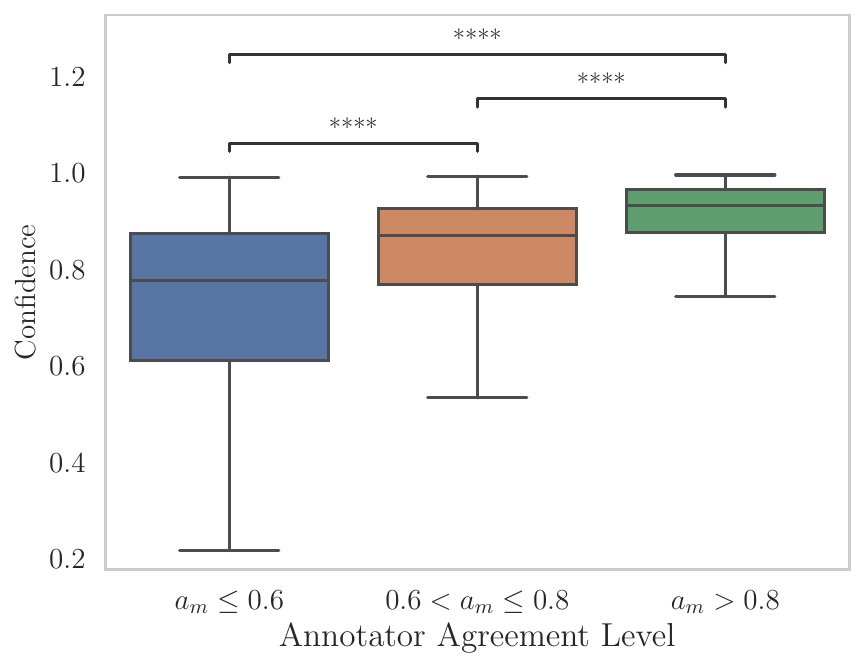}
    \includegraphics[width=0.32\linewidth]{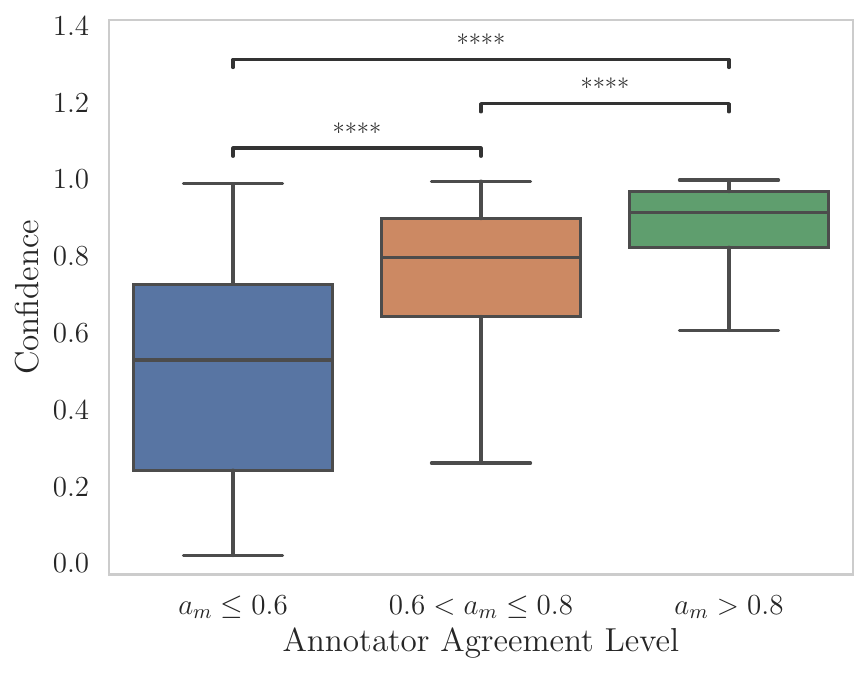}
        \includegraphics[width=0.32\linewidth]{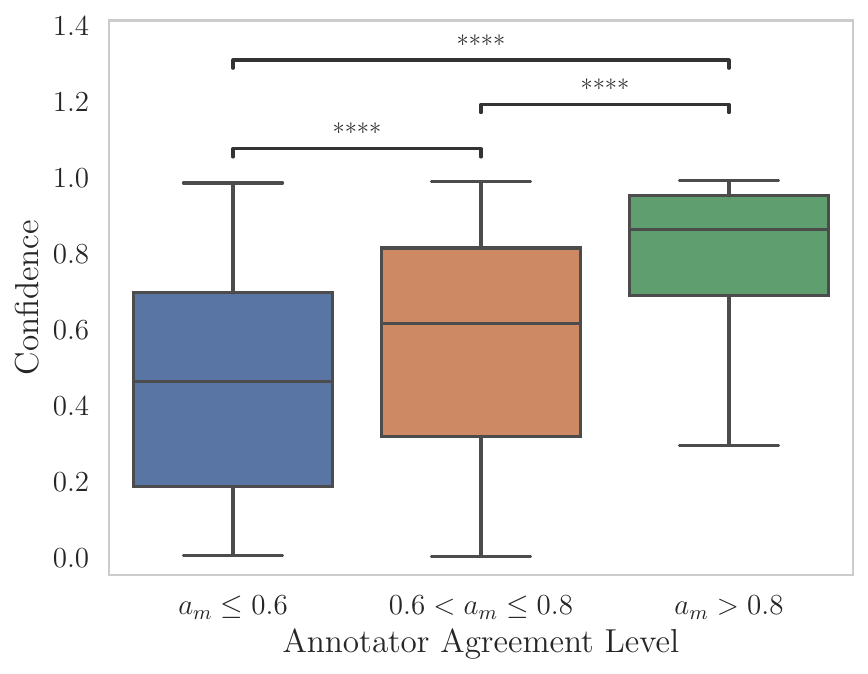}
    \caption{Boxplots illustrating the relationship between model confidence and annotator agreement level $(a_m)$ for Single-GT model trained on $\mathcal{D}_{\textsl{MDA}}$ (left), $\mathcal{D}_{\textsl{SI}}$ (center) and $\mathcal{D}_{\textsl{MHS}}$ (right). There is a clear correlation between model's confidence in predicting the ground truth label and the agreement between annotators (denoted as the fraction of annotators that agree on the majority vote on the x-axis). We further depict significant differences in confidence distribution across agreement levels using the Mann-Whitney-Wilcoxon test \cite{mcknight2010mann} with Statannotations \cite{florian_charlier_2022_7213391}. Notation includes \textbf{****} for $p <= 1.00e-04$.}
    \label{fig:single_gt_agr_factor_vs_conf}
\end{figure*}

\subsection{Results}
We calculate the training dynamics, confidence and variability to generate data cartography maps for the three datasets -- $\mathcal{D}_{\textsl{MDA}}$, $\mathcal{D}_{\textsl{SI}}$ and $\mathcal{D}_{\textsl{MHS}}$ -- as illustrated in Figure \ref{fig:single_gt_cartography}. Furthermore, we leverage training dynamics to evaluate the correlation between the model's confidence in predicting the gold label and the level of agreement among annotators for the gold label. This correlation is visually represented through boxplots in Figure \ref{fig:single_gt_agr_factor_vs_conf} for Single-GT model where gold label is the aggregated vote. Across all three datasets, we identify a robust correlation between model confidence and annotator agreement level. Notably, instances of higher disagreement among human annotators correspond to lower model confidence throughout training epochs when the model is trained on the majority vote. To quantify the observed correlation, we utilize Pearson correlation coefficient with the results shown in Table \ref{tab:single_gt_pearson_r_conf_vs_agr} where we see large correlation for all three datasets with the associated p-values being statistically significant.

It is worth noting that the model remains unaware of annotator agreement level information during training, as it is only trained on a single ground truth label for a text sample, which is the majority vote. Nevertheless, this external factor significantly influences the model's confidence, with instances of heightened disagreement among human annotators corresponding to a persistent trend of lower model confidence. Hence, a critical question arises: given the observed challenge where the model struggles to learn samples with high disagreement level exhibiting low confidence, can being exposed to multiple annotators' annotations enhance the model's learning capabilities on low confidence \textit{(hard-to-learn)} samples?



\begin{table}
    \centering
    \begin{tabular}{cccc}
    \hline
\textbf{Dataset} & $\mathcal{D}_{\textsl{MDA}}$ & $\mathcal{D}_{\textsl{SI}}$ & $\mathcal{D}_{\textsl{MHS}}$ \\
\textbf{Corr.}   & 0.44   & 0.37  & 0.45  \\
\hline
\end{tabular}
    \caption{Pearson correlation coefficients between model confidence on each sample and the corresponding annotator agreement level for Single-GT model trained on the three datasets. The reported values are statistically significant.
    }\label{tab:single_gt_pearson_r_conf_vs_agr}
\end{table}

\section{RQ2: Do Multi-GT models lead to better confidences on hard-to-learn samples?}
\label{sec:RQ2}

\subsection{Methods}
 \label{sec:multi}
For our Multi-GT model, we rely on DisCo (Distribution from Context), as introduced by \citet{weerasooriya-etal-2023-disagreement}, which is a neural model specifically designed for predicting labels assigned by individual annotators. Instead of considering items in isolation, this model takes annotator-item pairs as input and conducts inference by considering predictions from all annotators. The authors discover that incorporating annotator-specific modules into a classifier, as opposed to overlooking individual perspectives, leads to superior performance.

Following the DisCo model, in this study, the inputs consist of instance-annotation pairs $(x_m, y_{n,m})$, where $x_m$ represents the mth data item, and $y_{n,m}$ denotes the label annotator $n$ assigned to it. We adapt the calculation of confidence and variability based on the probabilities of gold annotation per instance-annotation. This approach yields multiple confidences per item, corresponding to the number of annotations available for that item.

\subsection{Results}


\begin{table}
    \centering
    \begin{tabular}{cccc}
    \hline
\textbf{Dataset} & $\mathcal{D}_{\textsl{MDA}}$ & $\mathcal{D}_{\textsl{SI}}$ & $\mathcal{D}_{\textsl{MHS}}$ \\
\textbf{Corr.}   & 0.46   & 0.44  & 0.51  \\
\hline
\end{tabular}
    \caption{Pearson correlation coefficients between model confidence on each sample and the corresponding annotator agreement level for DisCo trained on the three datasets. When computing the training dynamics for DisCo, the pair of text sample and annotator ID is distinct across the dataset, which results in multiple confidence values for each annotation for a text sample. The reported values are statistically significant.}
    \label{tab:DisCo_pearson_r_conf_vs_agr}
\end{table}

\begin{figure*}[ht!]
    \centering
    \includegraphics[width=0.32\linewidth]{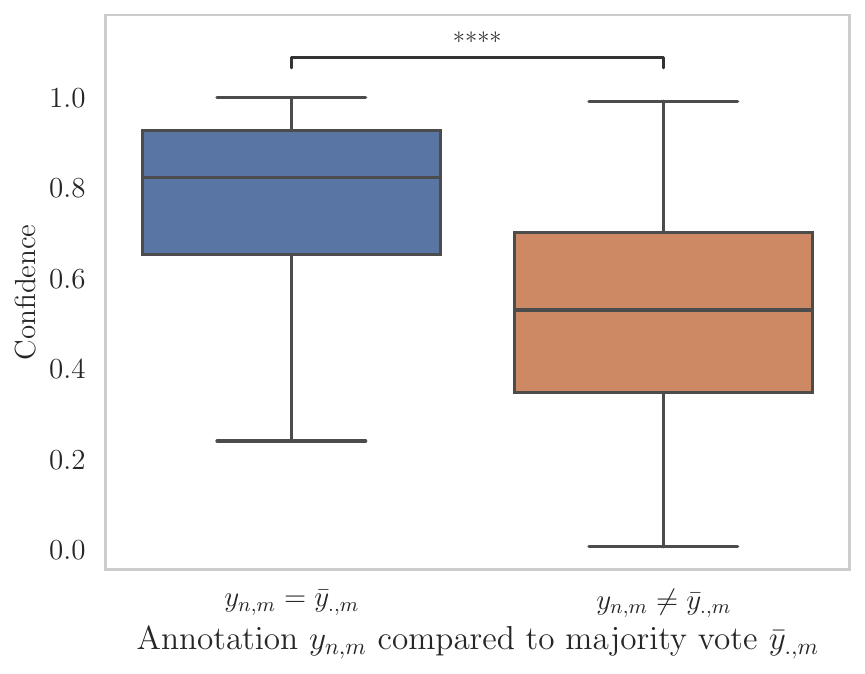}
    \includegraphics[width=0.32\linewidth]{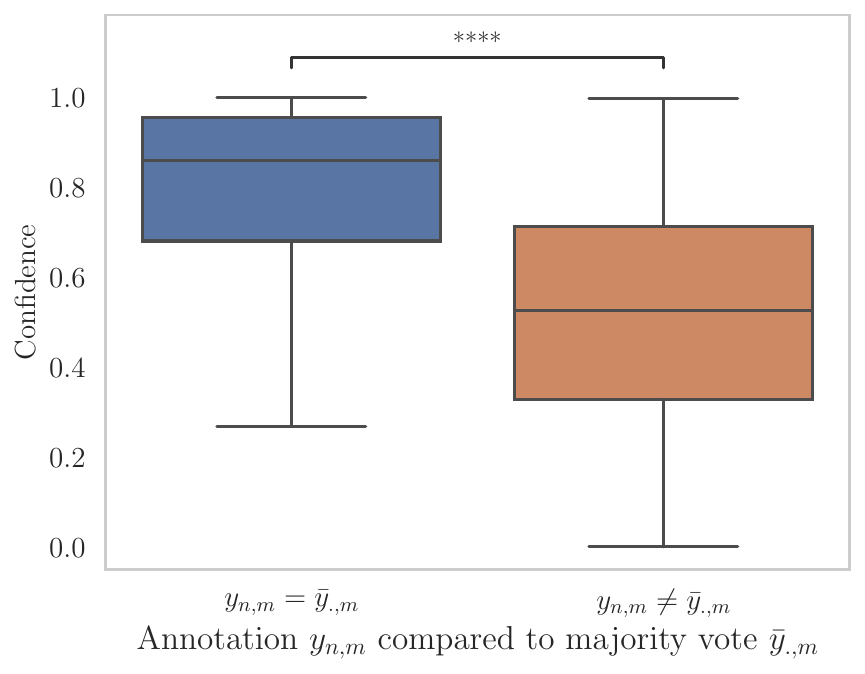}
    \includegraphics[width=0.32\linewidth]{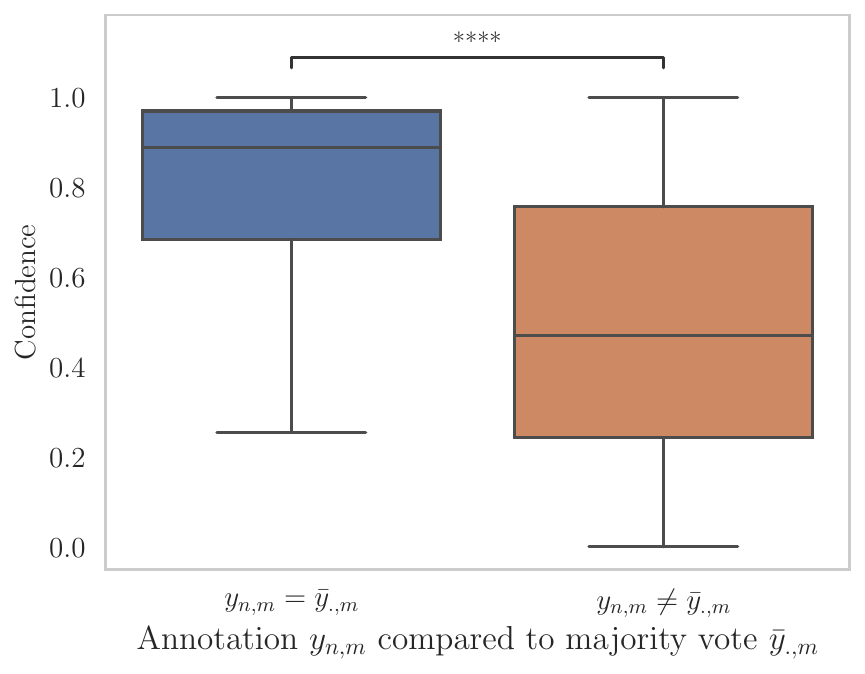}
    \caption{Boxplot illustrating the relationship between model confidence and whether the annotator's annotation ($y_{n,m}$) disagrees with the majority vote ($\bar{{y}}_{.,m}$) for DisCo trained on $\mathcal{D}_{\textsl{MDA}}$ (left), $\mathcal{D}_{\textsl{SI}}$ (center) and $\mathcal{D}_{\textsl{MHS}}$ (right). We see a clear correlation indicating higher confidence in the predicted label by the model when $y_{n,m} = \bar{{y}}_{.,m}$ and lower confidence when $y_{n,m} \neq \bar{{y}}_{.,m}$. We further depict significant differences in confidence distribution for $y_{n,m} = \bar{{y}}_{.,m}$ and $y_{n,m} \neq \bar{{y}}_{.,m}$ using the Mann-Whitney-Wilcoxon test \cite{mcknight2010mann} with Statannotations \cite{florian_charlier_2022_7213391}. Notation includes \textbf{****} for $p <= 1e-04$.}
    \label{fig:DisCo_disagree_vs_conf}
\end{figure*}

\begin{figure*}[ht!]
    \centering
    \includegraphics[width=0.32\linewidth]{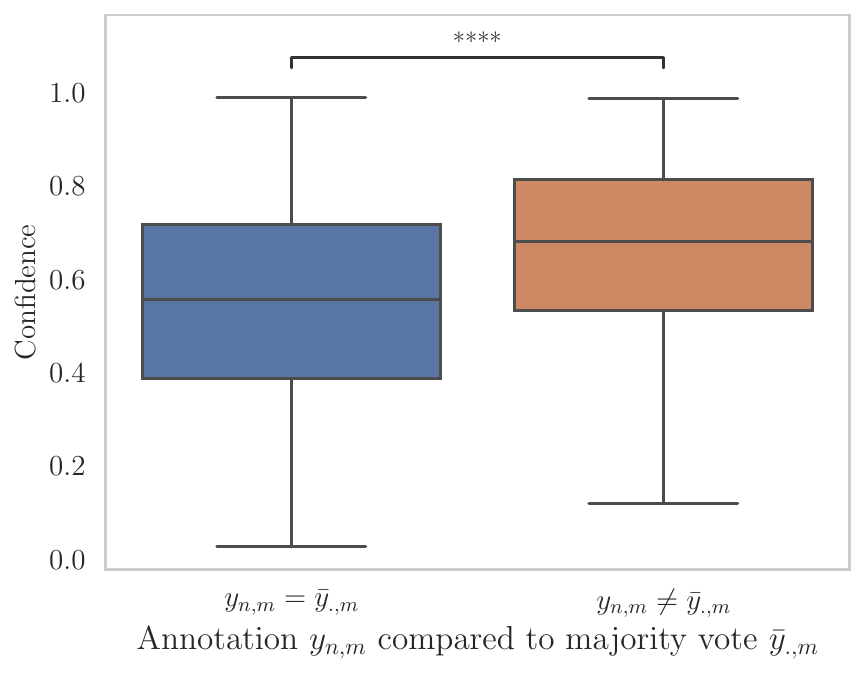}
    \includegraphics[width=0.32\linewidth]{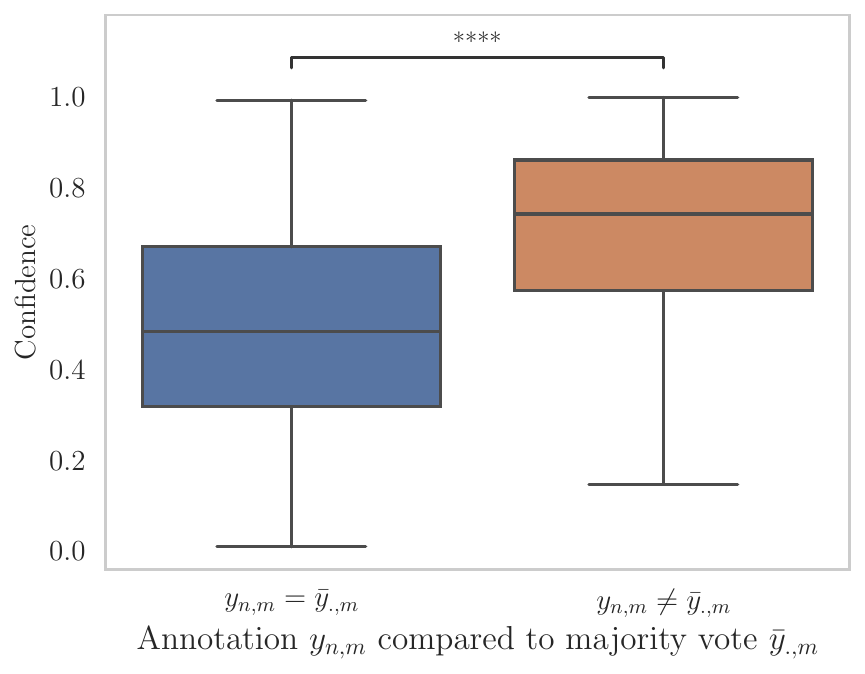}
    \includegraphics[width=0.32\linewidth]{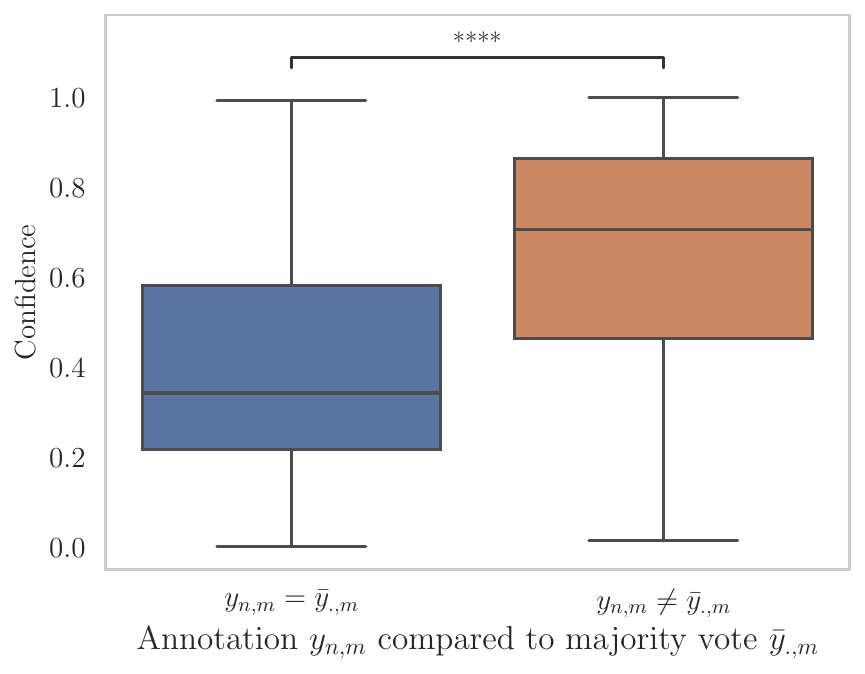}
    \caption{Boxplots illustrating the relationship between model confidence and whether the annotator's annotation ($y_{n,m}$) disagrees with the majority vote ($\bar{{y}}_{.,m}$) for DisCo trained on $\mathcal{D}_{\textsl{MDA}}$ (left), $\mathcal{D}_{\textsl{SI}}$ (center) and $\mathcal{D}_{\textsl{MHS}}$ (right) for DisCo only for the subset of samples where confidence is below 0.5 in Single-GT model. In contrast to the overall dataset presented in Figure \ref{fig:DisCo_disagree_vs_conf}, a reversed trend is observed, indicating higher confidence when $y_{n,m} \neq \bar{{y}}_{.,m}$ and lower confidence when $y_{n,m} = \bar{{y}}_{.,m}$. This highlights DisCo's ability to crucially learn from minority votes that are discarded for Single-GT model. We further depict significant differences in confidence distribution for $y_{n,m} = \bar{{y}}_{.,m}$ and $y_{n,m} \neq \bar{{y}}_{.,m}$ using the Mann-Whitney-Wilcoxon test \cite{mcknight2010mann} with Statannotations \cite{florian_charlier_2022_7213391}. Notation includes \textbf{****} for p <= 1.00e-04.}
    \label{fig:DisCo_disagree_vs_conf_for_low_conf_in_sgt}
\end{figure*}

As shown in the previous section, we employ training dynamics to assess the relationship between model confidence on annotations and the agreement level among annotators for a given text sample. We depict the relationship using Pearson correlation coefficient values in Table \ref{tab:DisCo_pearson_r_conf_vs_agr} with statistically significant p-values and the boxplots are illustrated in Appendinx \ref{sec:appendix}. It is important to note that for computing training dynamics for DisCo, the pair of text sample and annotator ID is unique across the dataset, hence, a text sample has multiple confidence values, one for each annotation for a text sample. We observe that consistent with the trend in models trained on a single ground truth label, heightened disagreement among annotators for a text sample correlates with reduced model confidence. We further check the model confidence distribution for annotations ($y_{n,m}$), grouped by whether they are equal to majority vote ($\bar{{y}}_{.,m}$) for all three datasets depicted in Figure \ref{fig:DisCo_disagree_vs_conf}. The results show a clear trend: samples with $y_{n,m} = \bar{{y}}_{.,m}$ yield a high-confidence distribution, while those with $y_{n,m} \neq \bar{{y}}_{.,m}$ result in a notably lower confidence distribution. Two factors may contribute to this observation: 1) the inclusion of noisy minority vote annotations, where the majority vote represents an objectively correct label; and 2) the architectural limitations of the model. Although the model is designed to learn multiple annotations for a given text sample depending on the annotator ID as input, it encounters challenges in confidently learning the minority vote annotation for the text. These results emphasize the significance of annotator agreement in understanding uncertainty in model predictions, which applies to both Single-GT model and DisCo, a Multi-GT model, where higher confidence aligns with increased agreement on annotations.

Additionally, to answer the question whether DisCo, a Multi-GT model, is able to demonstrate increased confidence levels in hard-to-learn instances for the Single-GT model, our investigation specifically targets text samples where Single-GT model exhibits low confidence (below 0.5). As illustrated in Figure \ref{fig:DisCo_disagree_vs_conf_for_low_conf_in_sgt}, a significant and consistent trend is observed across all three datasets. In this instance, samples with $y_{n,m} \neq \bar{{y}}_{.,m}$ show higher confidence compared to samples with $y_{n,m} = \bar{{y}}_{.,m}$. This contrasts with the relationship observed in the complete dataset boxplots in Figure \ref{fig:DisCo_disagree_vs_conf}, where model has higher confidence on samples with $y_{n,m} = \bar{{y}}_{.,m}$. This finding emphasizes a critical characteristic of DisCo, which can extract valuable information from annotations that are disregarded during the majority vote aggregation process. The Single-GT model never encounters this information and therefore cannot improve on challenging samples where the discarded annotation may be crucial due to mislabeled samples  \cite{swayamdipta-etal-2020-dataset} or the subjectivity of the text. 

We present a subset of the above group of samples with $y_{n,m} \neq \bar{{y}}_{.,m}$ in Table \ref{tab:mislabelled_examples} that have high confidence in DisCo (above 0.9, i.e. easy to learn) and low confidence in Single-GT model (below 0.5) for $\bar{{y}}_{.,m}$. Provided with the opportunity to learn the minority vote label $y_{n,m}$ for these samples, DisCo rather finds it easy to learn them and hence, leading to the conjecture that majority votes $\bar{{y}}_{.,m}$ are inaccurate. We provide an additional set of examples in Appendix \ref{sec:appendix} where the Single-GT model exhibits high confidence (above 0.5) for $\bar{{y}}_{.,m}$, while DisCo demonstrates extremely low confidence (below 0.1) for $y_{n,m}$ where $y_{n,m} \neq \bar{{y}}_{.,m}$. This observation suggests that, in these instances, minority votes $y{n,m}$ are deemed inaccurate.

\begin{figure*}[ht!]
    \centering
    \includegraphics[width=0.32\linewidth]{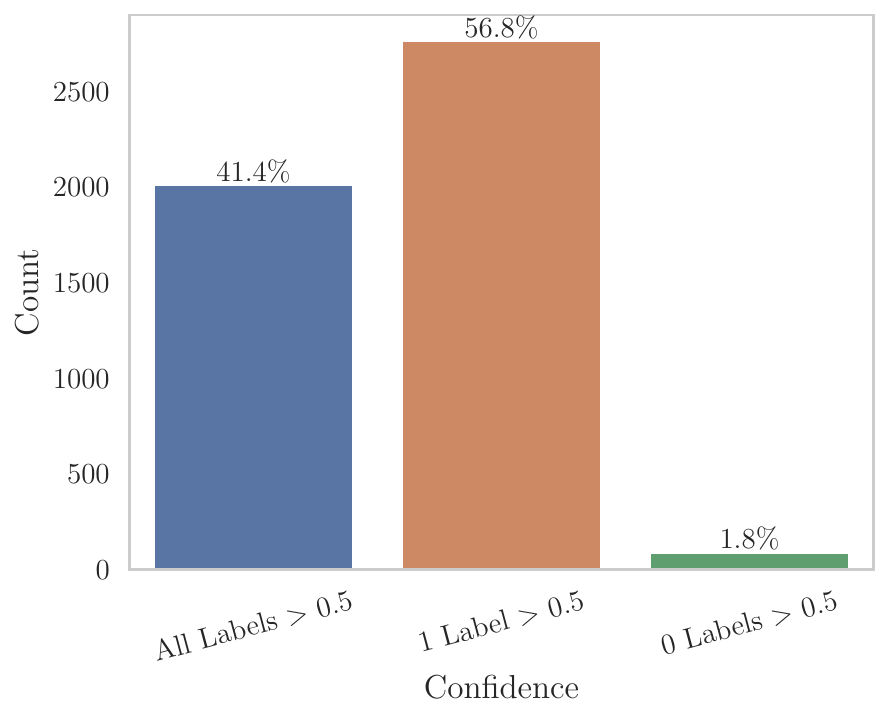}
    \includegraphics[width=0.32\linewidth]{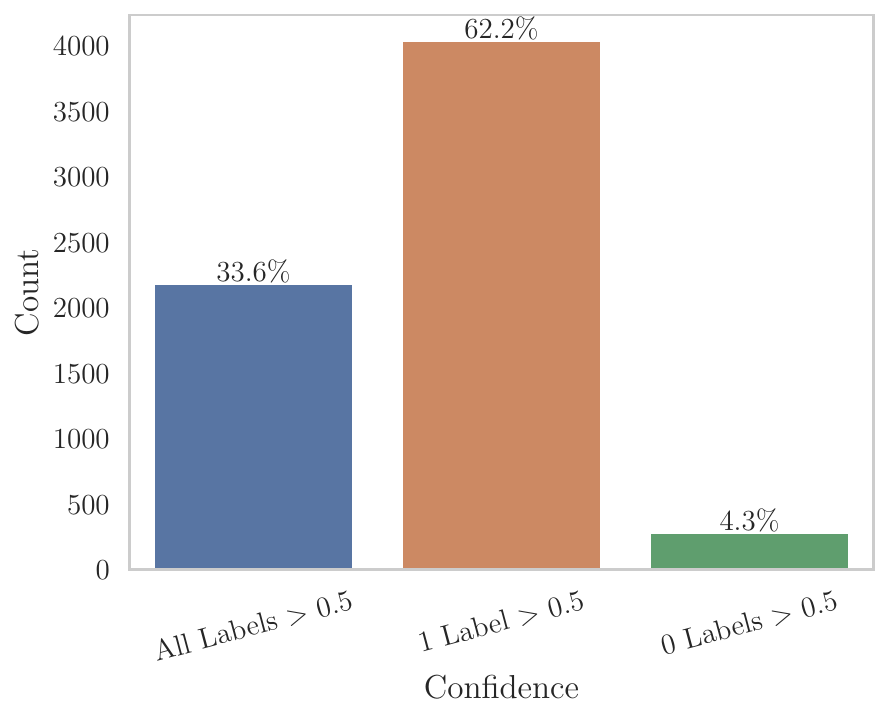}
    \includegraphics[width=0.32\linewidth]{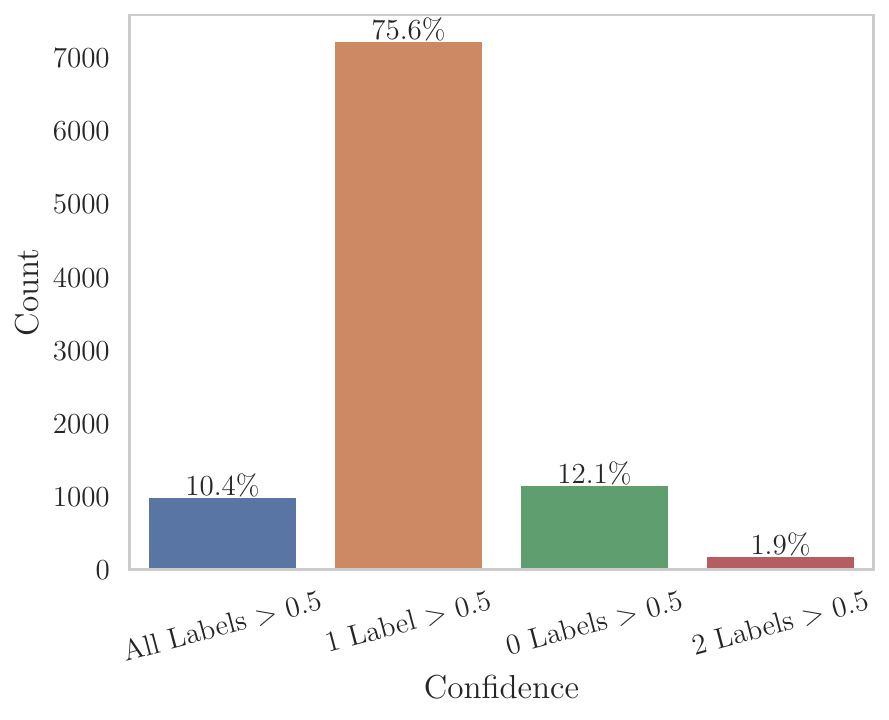}
    \caption{Number of samples with disagreement in the dataset where annotator agreement level is strictly below 1.0 and they are grouped by the number of different labels DisCo learns with high confidence (above 0.5) for these samples for the dataset $\mathcal{D}_{\textsl{MDA}}$ (left), $\mathcal{D}_{\textsl{SI}}$ (center) and $\mathcal{D}_{\textsl{MHS}}$ (right).}\label{fig:DisCo_conf_for_disagreement}
\end{figure*}

Further, to evaluate the model's capability to learn multiple annotator perspectives, we focus on samples with disagreement in the dataset where annotator agreement level is below 1.0, signifying disparate labels provided by different annotators for the same text. Effectively capturing diverse annotator perspectives entails the model's ability to accurately predict distinct labels for identical text inputs based on annotator input, showcasing its ability to learn varied perspectives encoded in the annotations. To illustrate this, in Figure \ref{fig:DisCo_conf_for_disagreement} we plot the count of samples with disagreement grouped by the number of different labels the model learns with high confidence (above 0.5). This visualization would help us assess whether the model is able to learn multiple labels for a text with high confidence, when the sole variation in input to the model lies in the annotator ID. Thus, it serves as an evaluation of its capability to learn different annotator viewpoints. For datasets $\mathcal{D}_{\textsl{MDA}}$ and $\mathcal{D}_{\textsl{SI}}$, with a binary classification task, although the model confidently learns a single label for over 50\% of the samples, there is still a notable subset of samples (All Labels > 0.5), where the model shows high confidence for both labels, indicating its ability to capture annotator perspectives. 

However, for $\mathcal{D}_{\textsl{MHS}}$, characterized by three labels, insights from Figure \ref{fig:DisCo_conf_for_disagreement} reveal that DisCo confidently learns only a single label for over 75\% of the samples, with approximately only 12\% samples where it confidently learns multiple labels. This underscores its challenge in capturing individual annotators' perspectives through their annotations. We attribute this difficulty to the notably low average number of annotations per annotator in $\mathcal{D}_{\textsl{MHS}}$ (below 20), as shown in Table \ref{tab:dataset}, in contrast to the other two datasets. The limited number of annotations per annotator presents an obstacle in effectively modeling an annotator's perspective. Therefore, we emphasize that accumulating a substantial number of annotations from each annotator is imperative for the effectiveness of DisCo.

Our analysis unveils key insights into model confidence and annotation dynamics. Examining the relationship between model confidence and annotator agreement levels for text samples, our findings echo those in Single-GT models, showing that heightened annotator disagreement aligns with decreased model confidence. In hard-to-learn instances for the Single-GT model, DisCo showcases increased confidence in samples with minority vote annotations, revealing its capacity to extract valuable insights from annotations typically overlooked in majority vote aggregation. Moreover, our investigation reveals that DisCo can effectively predict diverse labels for identical text inputs, especially in instances marked by disagreement, but it struggles in datasets with a limited number of annotations per annotator, emphasizing the necessity of accumulating a substantial number of annotations for DisCo's effectiveness. In essence, our findings underscore the critical importance of preserving multiple perspectives through annotations in subjective tasks and advocate for advancements in modeling approaches to achieve nuanced learning for broader representations.

\begin{table}[ht!]
\renewcommand{\arraystretch}{1.5}
\small
    \begin{adjustbox}{width=\columnwidth,center}
    \begin{tabular}{cp{4 cm}cc} \hline 
        \textbf{Dataset} & \multicolumn{1}{c}{\textbf{Text}} & \textbf{Majority Label} & \textbf{Minority Label}\\ \hline
        \multirow{3}{*}{$\mathcal{D}_{\textsl{MDA}}$} & START PUNCHING BACK !!! & offensive & not offensive\\ \cline{2-4} 
         & Y'ALL FORGOT THE WHITE DUDE THAT THE POLICE KILLED LAYING JUST LIKE GEORGE FLOYD THE CRIMINAL <url> & not offensive & offensive\\ \cline{2-4} 
         & His life mattered.  You however say whatever sounds good. It’s disgusting.  You insult this issue. & offensive & not offensive\\ \hline
        \multirow{3}{*}{$\mathcal{D}_{\textsl{SI}}$} & why are jews ' noses so big ? air is free alt : more gold to dig & not offensive & offensive\\ \cline{2-4} 
         & What did the Texas shooter say as he walked through the church? Pew pew pew! & not offensive & offensive\\ \cline{2-4} 
         & Still better that Nero's poetry. // cc @ChrisWarcraft & offensive & not offensive\\ \hline
        \multirow{3}{*}{$\mathcal{D}_{\textsl{MHS}}$} & Fucking fagot burn in hell & supportive & offensive\\ \cline{2-4} 
         & Your nigga don't suck toes or eat ass he a bitch. & supportive & offensive\\ \cline{2-4} 
         & At 7:19 is why I hate people, women especially look how mad she got just because the dog splashed her.. f*** you you stupid b**** either have fun or go kill yourself & neutral & offensive\\ \hline
    \end{tabular}
    \end{adjustbox}
    \caption{Examples from the three datasets $\mathcal{D}_{\textsl{MDA}}$, $\mathcal{D}_{\textsl{SI}}$ and $\mathcal{D}_{\textsl{MHS}}$ where Single-GT model has low confidence (below 0.5) for the Majority Label and DisCo has really high confidence (above 0.9) for the Minority Label. Following our best assessment, it appears that the majority label for this subset appears to be inaccurate, and the minority label emerges as the more suitable annotation.}
    \label{tab:mislabelled_examples}
\end{table}

\section{Conclusions}
This paper delves into an exploration of whether perspectivist classification models effectively harness valuable insights from instances identified as noisy through automated dataset evaluation techniques. Our investigation begins by examining how Single-GT models classify high-disagreement elements as noise. Subsequently, we shift our approach to Multi-GT models and observe a notable increase in confidence for minority votes for the same instances. This shift underscores the potential for richer and more nuanced understanding when leveraging multiple perspectives in the classification process. 

For future research directions, it is worth exploring model confidences for each annotator in the dataset in the context of the Multi-GT model. This investigation will enhance our understanding of the challenges faced by current models in learning annotator perspectives. Additionally, it is also worth exploring datasets like $\mathcal{D}_{\textsl{MHS}}$ featuring annotator demographic details and target demographic information for offensive text. Such datasets provide a chance to assess model confidences for both Single-GT and Multi-GT models across diverse demographic groups. This presents an opportunity to investigate the impact of preserving diverse perspectives through annotations in addressing societal biases within learned models.


\section*{Limitations}
Although we have carried out a comprehensive analysis, our study has certain limitations that warrant consideration. Firstly, the performance of Multi-GT models is dependent on the number of annotations per annotator, and a low number in some datasets may impact the representation of individual annotators. Secondly, the absence of raw annotations in many datasets limits a broader analysis of potential bias or noise. Additionally, variations in annotation instructions across datasets and differing levels of freedom for subjective interpretation among annotators introduce potential biases and inconsistencies that may affect comparison. Moreover, for Multi-GT models, this paper only considers DisCo, which requires an annotator ID to make the prediction. However, future research can explore the models that learn from the distribution of labels for each item. Furthermore, various approaches to defining annotators' label agreement, such as entropy and silhouette score \cite{mokhberian2022noise}, could be explored in forthcoming research. Finally, despite employing a Multi-GT approach, there is a possibility that the dataset items and annotators may have limitations as they may belong to a non-representative pool that does not encompass diverse societal perspectives. These limitations highlight the importance of cautious interpretation and generalization of our findings.




%
\section*{Ethical Considerations}
We employ Multi-GT models to capture diverse perspectives in the classifier. However, it's conceivable that the items or annotators within each collected dataset may be constrained in various ways, and the annotator pool may not accurately represent perspectives from the entire societal spectrum. Limitations could stem from factors such as an insufficient count of annotators from specific demographics in the pool or the presence of noisy annotations from certain annotators.

An additional ethical consideration in training Multi-GT models that capture the preferences of individual annotators is the issue of privacy and anonymity. It is crucial to ensure that annotators remain anonymized, and the process of learning and inferring their personal perspectives is conducted in a manner that avoids any potential misuse or harm.

\section*{Acknowledgments}
This work was funded in part by Defense Advanced Research Projects Agency (DARPA) and Army Research Office (ARO) under Contract No. W911NF-21-C-0002. We express gratitude to the anonymous reviewers for providing valuable feedback and offering suggestions for our project.

\bibliography{anthology,custom}

\appendix
\clearpage
\section{Supplemental Material}
\label{sec:appendix}

\subsection{Experimental Setup}

\begin{table}[ht!]
    \centering
    \begin{tabular}{ccc} \hline 
        \textbf{Dataset} & \textbf{F1 (Single-GT)} & \textbf{F1 (DisCo)}\\ \hline
        $\mathcal{D}_{\textsl{MDA}}$ & 0.78 & 0.78\\ 
        $\mathcal{D}_{\textsl{SI}}$ & 0.80 & 0.78\\ 
        $\mathcal{D}_{\textsl{MHS}}$ & 0.68 & 0.75\\ \hline 
    \end{tabular}
    \caption{F1(weighted) scores for Single-GT and DisCo trained on the three datasets.}
    \label{tab:f1_scores}
\end{table}

For our experiments we utilize pre-trained RoBERTa-Base as Single-GT model for fine tuning on $\mathcal{D}_{\textsl{MDA}}$, $\mathcal{D}_{\textsl{SI}}$ and $\mathcal{D}_{\textsl{MHS}}$. Both Single-GT and DisCo were trained for 5 epochs on each dataset to compute training dynamics values of confidence and variability. We report the F1 scores for Single-GT and DisCo models trained on $\mathcal{D}_{\textsl{MDA}}$, $\mathcal{D}_{\textsl{SI}}$ and $\mathcal{D}_{\textsl{MHS}}$ in Table \ref{tab:f1_scores} offering a summary of their performance to highlight convergence of models and reliability of predictions. 

\subsection{Additional Examples and Plots}
\begin{figure}[ht!]
    \centering
    \includegraphics[width=1.0\linewidth]{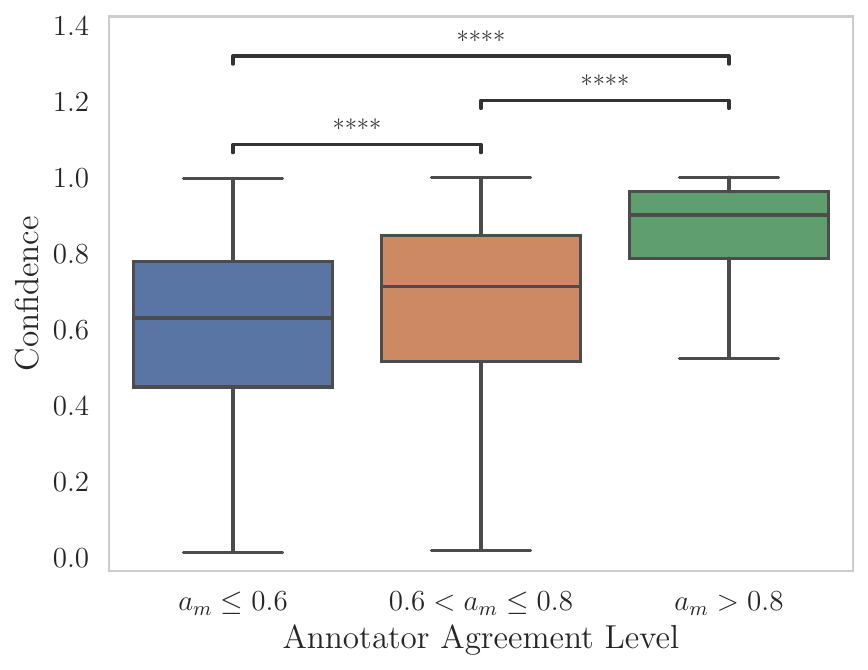}
    \includegraphics[width=1.0\linewidth]{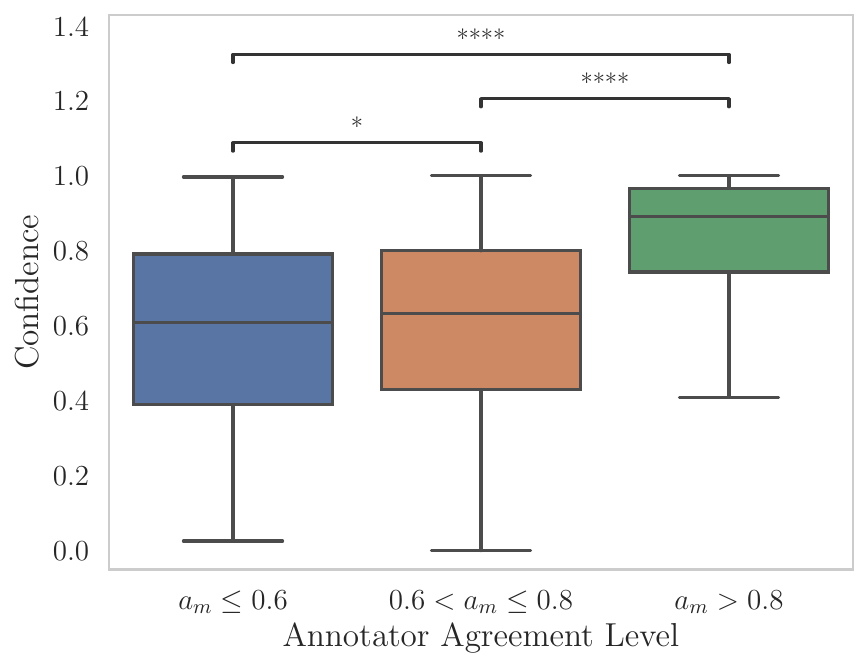}
        \includegraphics[width=1.0\linewidth]{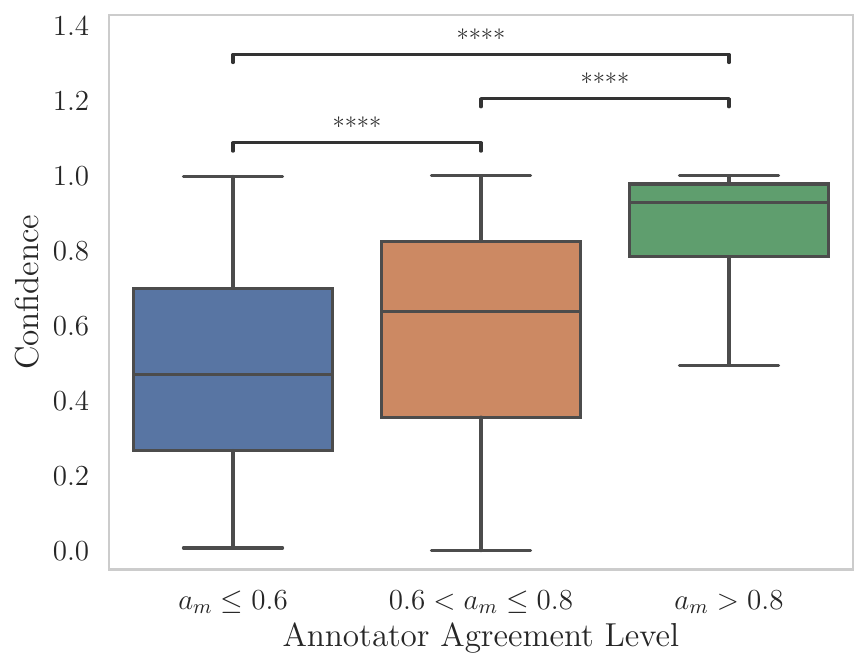}
    \caption{Boxplots illustrating the relationship between model confidence and annotator agreement level $(a_m)$ for DisCo trained on $\mathcal{D}_{\textsl{MDA}}$ (top), $\mathcal{D}_{\textsl{SI}}$ (center) and $\mathcal{D}_{\textsl{MHS}}$ (bottom). We see a clear correlation indicating higher confidence in the predicted label by the model with higher agreement between annotators (denoted as the fraction of annotators that agree on the majority vote on the x-axis).  We further depict significant differences in confidence distribution across agreement levels using the Mann-Whitney-Wilcoxon test \cite{mcknight2010mann} with Statannotations \cite{florian_charlier_2022_7213391}. Notation includes \textbf{*} for $1e-02 < p <= 5e-02$ and \textbf{****} for $p <= 1e-04$.}\label{fig:DisCo_agr_factor_vs_conf}
\end{figure}

\begin{table}[ht!]
\renewcommand{\arraystretch}{1.5}
\small
    \begin{adjustbox}{width=\columnwidth,center}
    \begin{tabular}{cp{4 cm}cc} \hline 
        \textbf{Dataset} & \multicolumn{1}{c}{\textbf{Text}} & \textbf{Majority Label} & \textbf{Minority Label}\\ \hline
        \multirow{3}{*}{$\mathcal{D}_{\textsl{MDA}}$} & Shocking! (Sorry, sarcasm doesn’t work well on Twitter) & not offensive & offensive\\ \cline{2-4} 
         & <user> Did he really die of COVID-19. ??? & not offensive & offensive\\ \cline{2-4} 
         & Fragility at its finest & not offensive & offensive\\ \hline
        \multirow{3}{*}{$\mathcal{D}_{\textsl{SI}}$} & Rate this Anti-Social Socially Anxious Hermit Loser with Long Hair (Pic) & offensive & not offensive\\ \cline{2-4} 
         & What's a pedophile's all you can eat buffet? The school bus & offensive & not offensive\\ \cline{2-4} 
         & But is it really that big a deal? & not offensive & offensive\\ \hline
        \multirow{3}{*}{$\mathcal{D}_{\textsl{MHS}}$} & @Cam\_Coles You don't know what I know. Name somebody colder, you monkey faced immigrant & offensive & supportive\\ \cline{2-4} 
         & When these niggas force me to say good morning URL & offensive & neutral\\ \cline{2-4} 
         & Lailat al Miraj mubarak to all Muslims.. I'm wishing for everyone's peace \& prosperity & supportive & neutral\\ \hline
    \end{tabular}
    \end{adjustbox}
    \caption{Examples from the three datasets $\mathcal{D}_{\textsl{MDA}}$, $\mathcal{D}_{\textsl{SI}}$ and $\mathcal{D}_{\textsl{MHS}}$ where Single-GT model has high confidence (above 0.5) for the Majority Label and DisCo has really low confidence (below 0.1) for the Minority Label. Following our best assessment, it appears that the minority label in this case appears to be inaccurate.}
    \label{tab:mislabelled_minority_examples}
\end{table}

\end{document}